# A PCA-BASED SUPER-RESOLUTION ALGORITHM FOR SHORT IMAGE SEQUENCES

*Carlos Miravet[a,b], Francisco B. Rodríguez[a]*

a) Escuela Politécnica Superior, Universidad Autónoma de Madrid, 28049 Madrid, SPAIN
b) SENER Ingeniería y Sistemas, S.A., Severo Ochoa 4, 28760 Tres Cantos (Madrid), SPAIN
carlos.miravet@sener.es, f.rodriguez@uam.es

## ABSTRACT

*In this paper, we present a novel, learning-based, two-step super-resolution (SR) algorithm well suited to solve the specially demanding problem of obtaining SR estimates from short image sequences. The first step, devoted to increase the sampling rate of the incoming images, is performed by fitting linear combinations of functions generated from principal components (PC) to reproduce locally the sparse projected image data, and using these models to estimate image values at nodes of the high-resolution grid. PCs were obtained from local image patches sampled at sub-pixel level, which were generated in turn from a database of high-resolution images by application of a physically realistic observation model. Continuity between local image models is enforced by minimizing an adequate functional in the space of model coefficients. The second step, dealing with restoration, is performed by a linear filter with coefficients learned to restore residual interpolation artifacts in addition to low-resolution blurring, providing an effective coupling between both steps of the method. Results on a demanding five-image scanned sequence of graphics and text are presented, showing the excellent performance of the proposed method compared to several state-of-the-art two-step and Bayesian Maximum a Posteriori SR algorithms.*

***Index Terms*—** superresolution, sequence processing, Principal Component Analysis, local image models, image restoration

## 1. INTRODUCTION

Sequence-based super-resolution (SR) addresses the problem of generating a high-resolution (HR) image given a set of low-resolution (LR) images of a scene, with at least sub-pixel shifts between frames. In the last decades, numerous approaches to solve the problem of SR have been proposed in the literature [1]. Between these, Bayesian *Maximum a Posteriori* (MAP) methods [2] have gained wide acceptance due to their robustness and flexibility to incorporate a priori constraints, although they are usually affected by high computational burdens. MAP methods attempt to solve the SR problem in a single step, by a complex, iterative procedure.

More computationally-efficient SR methods could be devised [1] by splitting the HR image reconstruction task into two simpler steps, which are carried out sequentially. The first-step is devoted to increase the sampling rate of the incoming LR images, by regression or scattered-point interpolation on sequence pixels, after projection onto the HR frame. The aim of the second step is to restore the degradations associated to the large integration area of the LR sequence pixels, combined with artifacts introduced by the previous processing scheme (residual registration and interpolation errors) or degradations linked to the image acquisition process (defocusing, atmospheric and motion blur, etc). These two-step methods generally trade some loss in quality or flexibility for a lower computational complexity that leads to faster execution times.

In previous papers [3-4], the authors presented a simple learning-based SR method capable of providing results of quality similar to standard MAP methods when operating on relatively long sequences of 25 images (corresponding to a 1 s sequence in the PAL video standard). A comparison of this method to a state-of-the art SR algorithm was reported in [5]. In this paper, it is presented a novel two-step method with a significantly improved interpolation step, which makes use of sub-pixel image statistics. This enables effective application of the SR method to very short image sequences, which is relevant not only from a theoretical, but also from a practical point of view. The fields of application in which fairly short sequences are typically only the ones available are numerous, ranging from neuroscience to astronomy, or applications related to surveillance and satellite image processing.

The paper is organized as follows. The observation model is briefly described in section 2. In section 3 are introduced the two steps of the proposed SR reconstruction method, together with the previous sub-pixel registration step. Experimental results are reported in section 4. Finally, section 5 concludes the paper.

## 2. OBSERVATION MODEL

Accurate estimates of the data provided by an imaging sensor are required at several stages of the learning phase of the algorithm proposed. Firstly, LR image patches sampled at sub-pixel resolution are required to derive the Principal Components (PC) used in the interpolation step of the algorithm. Secondly, the coefficients of the linear restoration filter used in the last step are fitted to minimize the mean square error (MSE) between HR reconstructed pixels, obtained by application of the interpolation-restoration paradigm on simulated LR sequences, and the goal HR values, derived by use of the observation model at the desired increase in resolution for the SR process.

Image patches at different levels of resolution are generated synthetically from a set of reference images, taken to

This work was supported by the Spanish Ministry of Education and Science under TIN 2007-65989 and CAM S-SEM-0255-2006, and by COINCIDENTE project DN8644, RESTAURAC.

represent a very high-resolution approximation of a continuous, degradation-free image. Using these images, the output of an imaging sensor of a given (lower) resolution could be simulated by adequate blurring, down-sampling and noise addition operations. Specifically, the following processing steps are sequentially applied to a reference image to obtain a simulated picture at a given resolution: a) filtering with the lens optical diffraction-limited transfer function (OTF); b) filtering with the Shannon empirical OTF formula, to account for symmetric aberrations in the optics; c) filtering with the detector OTF; d) down-sampling to the sensor resolution and e) addition of Gaussian noise. Details and mathematical expressions for the above mentioned OTF's could be found, for instance, in [6].

Figure 1 shows a comparison of the actual image (left) provided by a CCD camera when imaging the inner part of the 1951 USAF resolution test chart and the image (right) estimated by application of our observation model (with the actual camera parameters) to a digitized version of the resolution chart, obtained by use of a high-resolution flat-bed scanner. The visual agreement between both images is excellent, as the model succeeds in reproducing accurately the aliasing effects and the frequency-dependent lost of contrast in the bars, apparent in the CCD image.

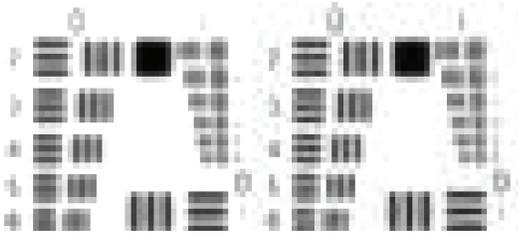

**Fig. 1.** *left*: image obtained from a CCD camera; *right*: estimated image, computed using the observation model

### 3. SUPER-RESOLUTION METHOD

The method proposed follows the pattern of two-step SR methods, with a first step of scattered point-interpolation of sequence pixels, followed by a restoration to correct for the LR pixel size and residual errors introduced by the interpolation procedure. Operation of the method requires previous application of a registration step, to obtain the geometric transform coefficients between input frames that will enable sequence pixels projection onto the HR image frame.

#### 3.1. Step 0: sub-pixel registration

This step is performed by application of a classical registration algorithm, capable of providing the geometric transformation coefficients between two images of the same scene with accuracy higher than the pixel size. Several motion models between frames of the input sequence were considered, ranging from pure translations to affine and planar projective transforms. Motion models up to affine were processed by suitable versions of the Irani-Peleg sub-pixel registration algorithm [7], with the extension to affine transformations presented in [3]. Fitting of the 8-dimensional planar projective model is performed by application of the Mann-Picard method [8].

#### 3.2. Step 1: scattered-point interpolation using PC

Using the geometric mappings between frames determined in the previous step, sequence pixels are projected onto the HR frame (aligned to one of the sequence frames), and the LR image values at the nodes of the HR grid are estimated from neighboring projected pixels. A diagram of the process is presented in figure 2.

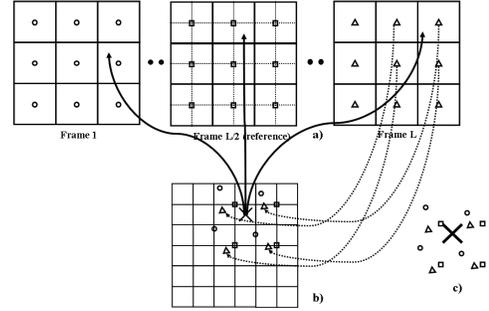

**Fig. 2.** sequence projection process: a) sequence pixels; b) projection onto the HR grid; c) data used to estimate a HR pixel (X)

One possible approach to solve this estimation problem is to use the projected data to fit local image models, which are subsequently applied to generate new image values at the HR grid nodes. In this approach, it stands as natural the use of models based on the principal components (PC) of the image at a local scale, as they will yield minimal representational error (in terms of mean squared error, MSE) for a given dimensionality of the model [9]. A preliminary study, based on the use of polynomial models, is reported in [10].

For this purpose, we have determined the PC of image patches of 64 x 64 pixels, spanning 4 x 4 pixels of a LR image. The densely sampled patches required have been generated by application of the observation model to samples of the reference image set, with appropriate blurring and downscaling parameters. Results obtained are in excellent agreement with those previously reported [11, 12] for the PC of local patches of natural images, in spite of the fact that, in our case, reference images are constituted by HR satellite images of urban scenes [3], and sub-pixel data is generated using an observation model.

Inside an image patch of size 4 x 4 LR pixels, the image structure at sub-pixel level has been modeled as a linear combination of continuous functions obtained by smooth bi-cubic interpolation of the first 60 PC. Coefficients of the linear combination ($a_j$) have been determined by minimization of the MSE between model predictions and the projected sequence pixel values in the neighborhood:

$$E = \frac{1}{2}\sum_i \left( z_i - \sum_j a_j f_{PCj}(x_i, y_i) \right)^2, \qquad (1)$$

where E is the model prediction error, $z_i$ the $i$ pixel value with projected coordinates ($x_i$, $y_i$) and $f_{PCj}$ is the continuous function obtained by interpolation of the $j^{th}$ PC.

Solution of this general linear least square problem requires a matrix inversion which has been performed using singular value decomposition, SVD [13]. The use of SVD improves robustness when the problem is close to singular, due, for instance, to an inadequate model order selection in an image patch, or induced by a large image noise level.

Local model representations centered in the nodes of a grid with a spacing of 2 LR pixel size have been computed, covering the complete extension of the HR image to be reconstructed. Taking into account the 4 x 4 pixel size of the local representation, this provides an area overlap of 1/2 between neighboring models, improving continuity at the model boundaries. Residual continuity errors are reduced to a negligible level by application of a conjugate gradient descent optimization process [13] on the space of PC coefficients.

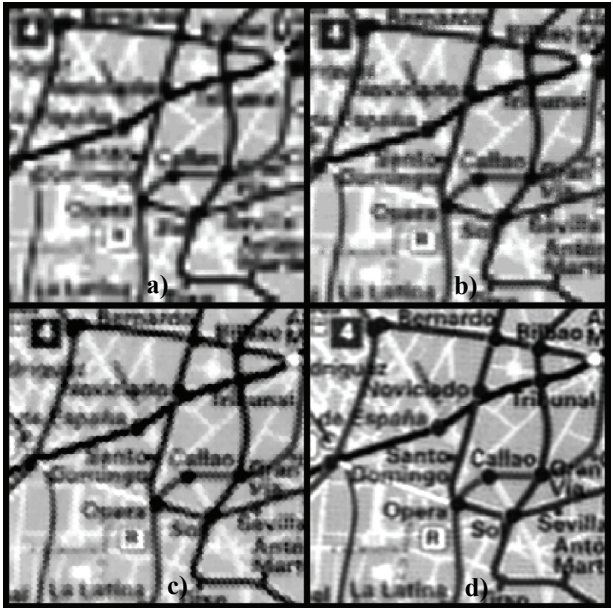

**Fig. 3**. Results of the interpolation step for two-step methods (see text)

### 3.3. Step 2: learning-based image restoration

Following the approach described in [3], the restoration step is performed by application of a rotationally symmetric linear filter in the spatial domain. Filter coefficients are determined by minimizing MSE between restoration of step one interpolated images and goal HR image values. This process implicitly considers residual degradations caused by interpolation, effectively coupling both steps of the super-resolution method.

### 4. EXPERIMENTAL RESULTS

The SR images obtained by application of the proposed SR method have been compared with excellent results to those provided by several state-of-the-art SR algorithms, including one and two-step HR image reconstruction procedures, on a variety of acquired sequences of different kind. Due to space constraints, we present here results on a single sequence, consisting of five frames depicting a map of Madrid's suburban rail network, digitized using a flat-bed scanner operating at low resolution (200 dpi), with small, uncontrolled specimen motions between acquisitions. This sequence is particularly suitable to spot the variations in performance of the different methods, due to the presence of a large amount of details of different type (points, lines, text), all highly-contrasted in the original. The results presented for methods other than those developed by the authors were obtained using the MDSP Resolution Enhancement Software [14], which integrates several state-of-the-art SR algorithms. Method's internal parameters, in all cases, have been set to the default values included in the SW.

In figure 3 are presented the results of the first step of the reconstruction (i.e. interpolation) for some of the two-step methods under comparison. In panel a) is presented, for reference, the results obtained with a cubic spline interpolation of a single frame. As is apparent, text remains clearly under-resolved and jaggedness, due to aliasing, is notorious in the graphic lines. All SR interpolation methods tested showed significant improvements with respect to this control image. In b) are presented the results obtained with the 'Shift and Add' (S&A) method, as described in [15], with an additional preprocessing outlier detection/removal algorithm using a bilateral filter [16]. Despite the increase in resolution with respect to the cubic spline interpolated image, jaggedness in diagonal lines is still apparent, text is still not fully resolved and is affected at some degree by impulse noise. Results with a median S&A method [17] are for this sequence visually similar to those displayed in b). In panel c) are presented the results obtained with the SR interpolation method described in [3]. As expected, the use of distance-dependent interpolation functions with sparse projected sequence data produces a blocky appearance in the details, with small-scale discontinuities in the graphic lines. Finally, the results of the proposed PCA-based interpolation are displayed in d). As could be observed, the method provides excellent results both in terms of resolution increase and absence of image artifacts (line jaggedness, regularity of details). Residual continuity errors at boundaries of local PC models are undetectable by visual inspection, event after the restoration performed in the second step of the reconstruction method.

In figure 4 are presented the results of the global HR reconstruction process for our method, together with those of several state-of-the-art methods being compared. In a) are presented the results of applying the bilateral S&A method followed by a Lucy deconvolution. This restoration step increases here significantly the irregularities present in the interpolation result, with noticeable artifacts in lines and text. In spite of this action, text remains to some extent under-resolved. Better results in terms of details resolution, but still with a somewhat high level of artifacts, are obtained by application of an S&A method with iterative deblurring, panel b). This method [17] applies the S&A step, performing afterwards the interpolation and deblurring operations in a single step, using a $L2$ (Tikhonov) regularization term. In c) and d) are presented the results of applying two state-of-the-art maximum likelihood (ML) type methods using regularization terms based, respectively, in L1 (Bilateral TV) and L2 (Tikhonov) measures. The L1 method [17], panel c), provides here noticeably clean results in low-complexity areas, at the cost of a perceptible degradation of small details, especially text, which present a washed-out appearance. Better results have been obtained with the L2 method [18], in panel d), but some irregularities in lines and text are still noticeable. In e) are presented the results obtained with the two-step SR method described in [3]. The quality of the restoration step, consisting of an optimal linear filtering process with pre-defined coefficients, is affected by the artifacts introduced by the interpolation step, with noticeable distortions in some texts, even if lines have more regular appearance than in other methods. Finally, in f) are presented the results of the proposed method, with perhaps the most successful balance between resolution increase and regularity of results.

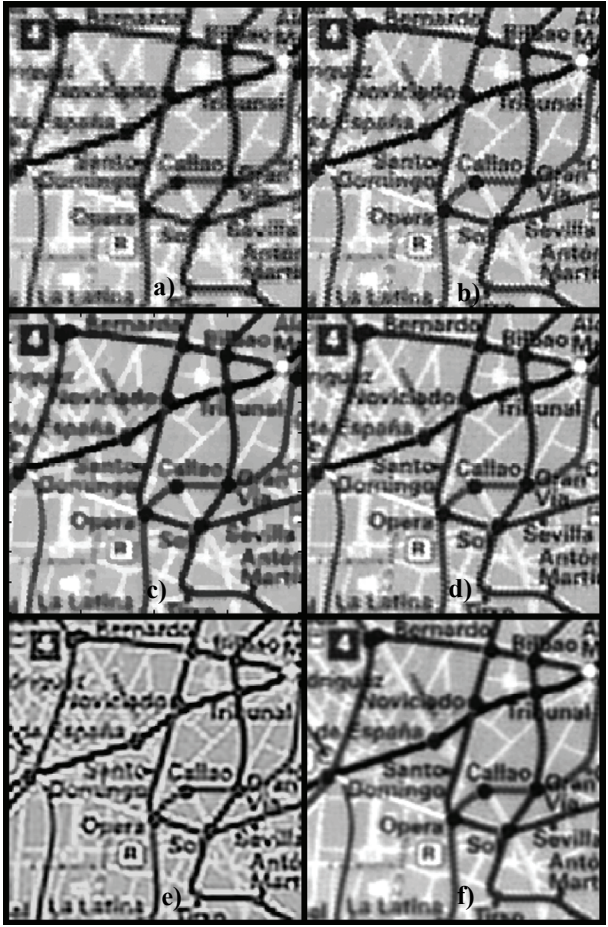

**Fig. 4**. SR results for different methods (see text)

## 5. CONCLUSIONS

In this paper, we presented a novel two-step super-resolution algorithm, with a robust scattered-point interpolation step well suited to solve the specially demanding problem of obtaining SR estimates from short image sequences. The results provided by the method are excellent in comparison not only to other two-step SR methods but also to state-of-the-art MAP-type algorithms. This indicates that the potential shortcomings of two step methods, due to the decoupling of interpolation and restoration operations, may be overcome by an adequate, data-driven definition of the operations to be performed in both steps of the method. In view of the results obtained, this might remain valid even for specially demanding scenarios, such as the SR reconstruction from short image sequences.

## 6. REFERENCES

[1] S.C. Park, M.K. Park and M.G. Kang, "Super-resolution image reconstruction: a technical overview", *IEEE Sig. Process. Mag*. 20 (3), 21–36, 2003.

[2] R.R. Schultz and R.L. Stevenson, "Extraction of high-resolution frames from video sequences", *IEEE Trans. Img. Process*., 5 (6), 996–1011, 1996.

[3] C. Miravet and F.B. Rodríguez, "A two-step neural-network based algorithm for fast image super-resolution", *Img. Vision Comput.,* 25 (9), pp. 1449–1473, 2007.

[4] C. Miravet and F.B. Rodriguez, "A hybrid MLP-PNN architecture for fast image superresolution", *Lect. Not. Comput. Sci.*, 2714, 417-425 , 2003.

[5] G. Cristóbal, E. Gil, F. Sroubek, J. Flusser, C. Miravet and F.B. Rodríguez, "Superresolution imaging: a survey of current techniques", *SPIE*, 7074, San Diego, CA, 2008.

[6] G.C. Holst, *Electro-Optical Imaging System Performance*, 4$^{th}$ Ed., SPIE Press, JCD Publishing, Washington, 2006.

[7] M. Irani and S. Peleg, "Improving resolution by image registration", *CVGIP: Graph. Model. Img. Process*. 53 (3), pp. 231–239, 1991.

[8] S. Mann and R.W. Picard, "Video Orbits of the Projective Group: "A Simple Approach to Featureless Estimation of Parameters", *IEEE Trans. Img. Process.*, 6 (9), pp 1281-1295, 1997.

[9] A. Bishop, *Neural Networks for Pattern Recognition*. Oxford University Press, Oxford, 1995.

[10] C. Miravet and F.B. Rodríguez. "Accurate and Robust Image Superresolution by Neural Processing of Local Image Representations". *Lect. Notes Comput. Sci*, 3696, pp. 499-505, 2005.

[11] P.J.B. Hancock, R.J. Baddeley and L.S. Smith, "The Principal Components of Natural Images", *Network: Comput. Neur. Syst.*, 3 (1), pp. 61-70, 1992.

[12] G. Heidemann, "The Principal Components of Natural Images Revisited", *IEEE Trans. Patt. An. Mach. Intell.*, 28 (5), pp. 822-826, 2006.

[13] W.H. Press, S.A. Teukolsky, W.T. Vetterling and B.P. Flannery, *Numerical Recipes in C: The Art of Scientific Computing*, 2$^{nd}$ Ed., Cambridge University Press, 1992.

[14] available at URL: http://users.soe.ucsc.edu/~milanfar/software/superresolution.html

[15] M. Elad and Y. Hel-Or, "A fast super-resolution reconstruction algorithm for pure translational motion and common space invariant blur", *IEEE Trans. Img. Process.*, 10 (8), pp. 1187-1193, 2001.

[16] S. Farsiu, D. Robinson, M. Elad and P. Milanfar, "Robust shift and add approach to super resolution", *Proc. of the 2003 SPIE Conf. Appl. Dig. Sig. Img. Process.*, pp. 121-130, 2003.

[17] S. Farsiu, D. Robinson, M. Elad and P. Milanfar, "Fast and robust multi-frame super-resolution", *IEEE Trans. Img. Process.* , 13 (10), pp. 1327-1344, 2004.

[18] M. Elad and A. Feuer, "Restoration of single super-resolution image from several blurred, noisy and down-sampled measured images", *IEEE Trans. Img. Process.*, 6 (12), pp. 1646-1658, 1997.